\documentclass[10pt,twocolumn,letterpaper]{article}

\usepackage{iccv}
\usepackage{times}
\usepackage{epsfig}
\usepackage{graphicx}
\usepackage{amsmath}
\usepackage{amssymb}
\usepackage{bbm}
\usepackage{booktabs}
\usepackage{multirow}
\usepackage{multicol}
\usepackage{bm}

\usepackage[pagebackref=true,breaklinks=true,letterpaper=true,colorlinks,bookmarks=false]{hyperref}

\iccvfinalcopy 

\def\iccvPaperID{7401} 

\ificcvfinal\fi

\begin{document}

\title{GloTSFormer: Global Video Text Spotting Transformer}

\newcommand{\model}[1]{GloTSFormer}

\author{Han Wang\\
ByteDance Inc.\\
\small{toameek@gmail.com}
\and
Yanjie Wang\\
ByteDance Inc.\\
\small{princewang1994@gmail.com}
\and
Yang Li\\
ByteDance Inc.\\
\small{ylijyangr@gmail.com}
\and
Can Huang\\
ByteDance Inc.\\
\small{can.huang@bytedance.com}
}

\maketitle
\ificcvfinal\thispagestyle{empty}\fi

\begin{abstract}
   Video Text Spotting (VTS) is a fundamental visual task that aims to predict the trajectories and content of texts in a video. Previous works usually conduct local associations and apply IoU-based distance and complex post-processing procedures to boost performance, ignoring the abundant temporal information and the morphological characteristics in VTS. In this paper, we propose a novel Global Video Text Spotting Transformer \model{} to model the tracking problem as global associations and utilize the Gaussian Wasserstein distance to guide the morphological correlation between frames. Our main contributions can be summarized as three folds. 1). We propose a Transformer-based global tracking method \model{} for VTS and associate multiple frames simultaneously. 2). We introduce a Wasserstein distance-based method to conduct positional associations between frames. 3). We conduct extensive experiments on public datasets. On the ICDAR2015 video dataset, \model{} achieves \textbf{56.0} MOTA with \textbf{4.6} absolute improvement compared with the previous SOTA method and outperforms the previous Transformer-based method by a significant \textbf{8.3} MOTA.
\end{abstract}

\section{Introduction}
\label{sec:intro}
\begin{figure}[!t]
    \centering
    \includegraphics[width=0.450\textwidth]{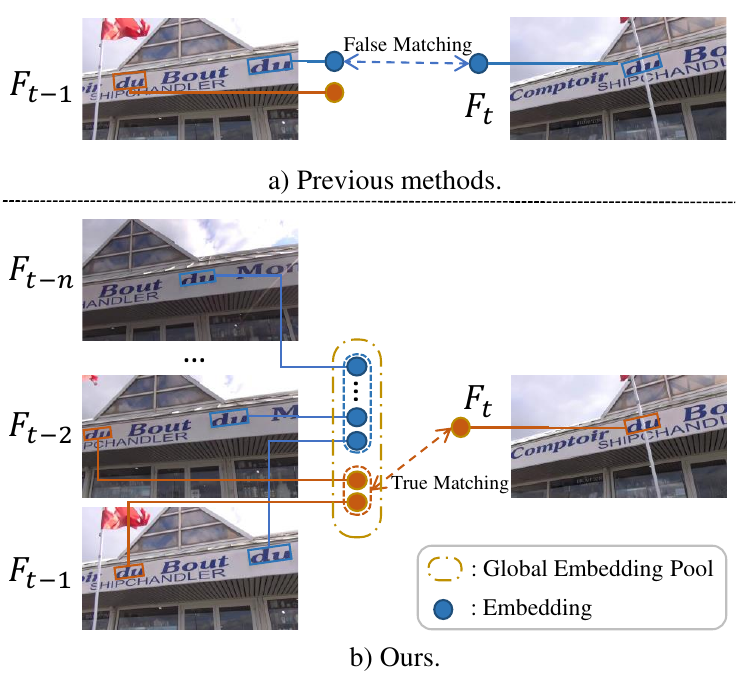}
    \caption{Motivation. Previous works usually conduct local associations and easily fail in scenes with interference (\textit{e.g.,} identical texts). To solve the problems, we introduce global associations to utilize temporal information to make our method more robust towards such scenes.}
    \label{fig:motivation}
\end{figure}
Video Text Spotting is an essential topic in computer vision, which facilitates video understanding, video retrieval, and video captioning. By simultaneously carrying out detection, recognition, and tracking, VTS can locate and recognize the texts in each frame and build trajectories through time. \\
\indent As shown in Fig. \ref{fig:motivation}, existing approaches~\cite{CoText, AJOU, CTPN, Free, STCM, SAVTD} have achieved success by modeling tracking as a bilateral matching problem between two local frames and applying an \textit{IoU}-based distance and cascaded post-processing procedures to build trajectories, similar to pipelines in Multi-Object Tracking (MOT). For instance, some works \cite{Free, CoText, SAVTD, online} apply metric learning \cite{metriclearning} methods to match the texts in two adjacent frames, measuring appearance similarity via cosine distance, and use cascaded matching based on \textit{IoU} distance just as \cite{FairMOT, JDE} for post-processing. However, we argue that there are two limitations. First, there exists abundant temporal information in VTS as most videos 
have a high frame rate and strong context coherence, which can be used to alleviate image deterioration problems (\textit{i.e.}, motion blur, lighting change, deformation, and similar instances). Second, the morphological information in VTS is understudied because the pipeline is borrowed from MOT without proper consideration of morphology. Different from frequent shape changes caused by limb deformation in MOT, the shapes of texts in VTS are much more stable within a short time, which can be utilized to boost the performance of tracking as an important feature. 
Thus, how to explicitly utilize temporal information and properly conduct morphological correlation remains a question.\\
\indent In this paper, we propose a novel model \model{} with global associations to explicitly use temporal information and a shape-aware distance to measure morphological similarity. We modify the detector YOLOX~\cite{YOLOX} to detect texts as polygons in each frame. The tracking embeddings are extracted by Rotated RoIAlign~\cite{FOTS} and supervised by recognition loss to obtain semantic awareness. To utilize temporal information, a global embedding pool is maintained during the whole inference process to hold the historical tracking embeddings and trajectory information. Then a Transformer-based architecture is proposed to access long-range temporal associations by conducting associations between texts in the current frame and texts in the global embedding pool for each frame. We also introduce a Wasserstein distance-based~\cite{GWD} method as the positional measurement, which takes both location and morphology into account. \\
\indent To prove the effectiveness of the proposed method, we conduct extensive experiments on several datasets and achieve state-of-the-art performance. On ICDAR2015 video~\cite{ICDAR2015} dataset, our \model{} obtains \textbf{56.0} MOTA on the test split, with \textbf{4.6} absolute improvement compared with the previous SOTA method~\cite{CoText}, and outperforms the previous Transformer-based method~\cite{TransDETR} by \textbf{8.3} MOTA. On the ICDAR2013~\cite{ICDAR2013} video and Minetto~\cite{Minetto} datasets, our \model{} also reaches leading performance. Our \model{} can run at around 20 FPS and the global association procedure takes 3.6 ms per frame on a single Tesla V100 GPU.\\




\section{Related work}
\subsection{Scene Text Detection}
Different from object detection, Scene Text Detection aims to detect arbitrarily shaped texts in images. Benefiting from the development of object detection, \cite{CTPN, Deeptext} succeed in horizontal text detection by adopting similar methods. Based on FCN \cite{FCN}, EAST \cite{EAST} is proposed to detect texts with different angles. PSENet \cite{PSENet} and PAN/PAN++ \cite{Pan, Panpp} adopt kernel-based methods and apply post-processing procedures to produce the final detecting results. 

\subsection{Multi-Object Tracking}
Multi-Object Tracking aims to predict the coordinates of each object in each frame. Most existing methods \cite{SORT, DeepSORT, CenterTrack, FairMOT, JDE} model the tracking task as a bilateral matching problem between instances in two adjacent frames. \cite{DeepSORT} adopts separate detection and tracking networks and a tracking-by-detection pipeline, with an \textit{IoU}-based positional distance and cascaded matching procedures. To simplify the pipeline, \cite{JDE, FairMOT} introduce a joint-detection-and-tracking protocol, which combines both detection and tracking in a single network. However, they highly rely on complex post-processing procedures and require many handcrafted hyper-parameters. Recently, some works \cite{TransTrack, MOTR} model tracking as a query problem, treating different trajectories as different queries and decoding the corresponding coordinates with a Transformer-based architecture. Though a more precise pipeline, these approaches usually fail in crowded scenes due to the absence of explicit positional awareness. 

\subsection{Video Object Detection}
Video Object Detection (VOD) aims to boost the detection performance by aggregating context features. Attention blocks are widely used in \cite{MEGA, SELSA, RDN, LongRange, TransVOD, PTSEFormer, FGFA} to conduct correlations between reference images and the current image, achieving awareness of long-range temporal information. Based on a two-stage detector, \cite{MEGA, SELSA, RDN, LongRange, FGFA} aggregate features after RoIs \cite{fastrcnn, maskrcnn} to gain enhanced features. In particular, \cite{MEGA} designs a hierarchical structure to aggregate local and global features. Based on Transformer, \cite{PTSEFormer, TransVOD} aggregate queries through time, also achieving temporal awareness.

\subsection{Video Text Tracking and Video Text Spotting}
Given a video clip, Video Text Tracking (VTT) aims to predict the coordinates of each text in each frame and Video Text Spotting further requires recognition results. Existing methods \cite{SVRep, CoText, TransDETR, TransVTSpotter, Free, SAVTD} succeed in most common scenes by conducting local associations. For example, a typical structure consists of a backbone, a detector and an RoI to extract instance-level features. The appearance similarity between instances is measured by a pairwise distance (\textit{e.g.,} cosine distance). The positional association score is calculated by the \textit{IoU} between instances in frames. A cascaded post-processing is applied to fully use the appearance similarity and positional association. Motivated by \cite{MOTR, TransTrack}, some models~\cite{TransDETR, TransVTSpotter} directly apply Transformer-based architectures to VTS, with an extra network for recognition. However, lacking \textit{IoU}-based post-processing procedures and utilization of temporal information, these methods struggle in many difficult VTS situations (\textit{e.g.,} crowded scenes, fast movement, lighting change).

\section{Methods}
\subsection{Overview}
\begin{figure*}[!t]
    \centering
    \includegraphics[width=1.0\linewidth]{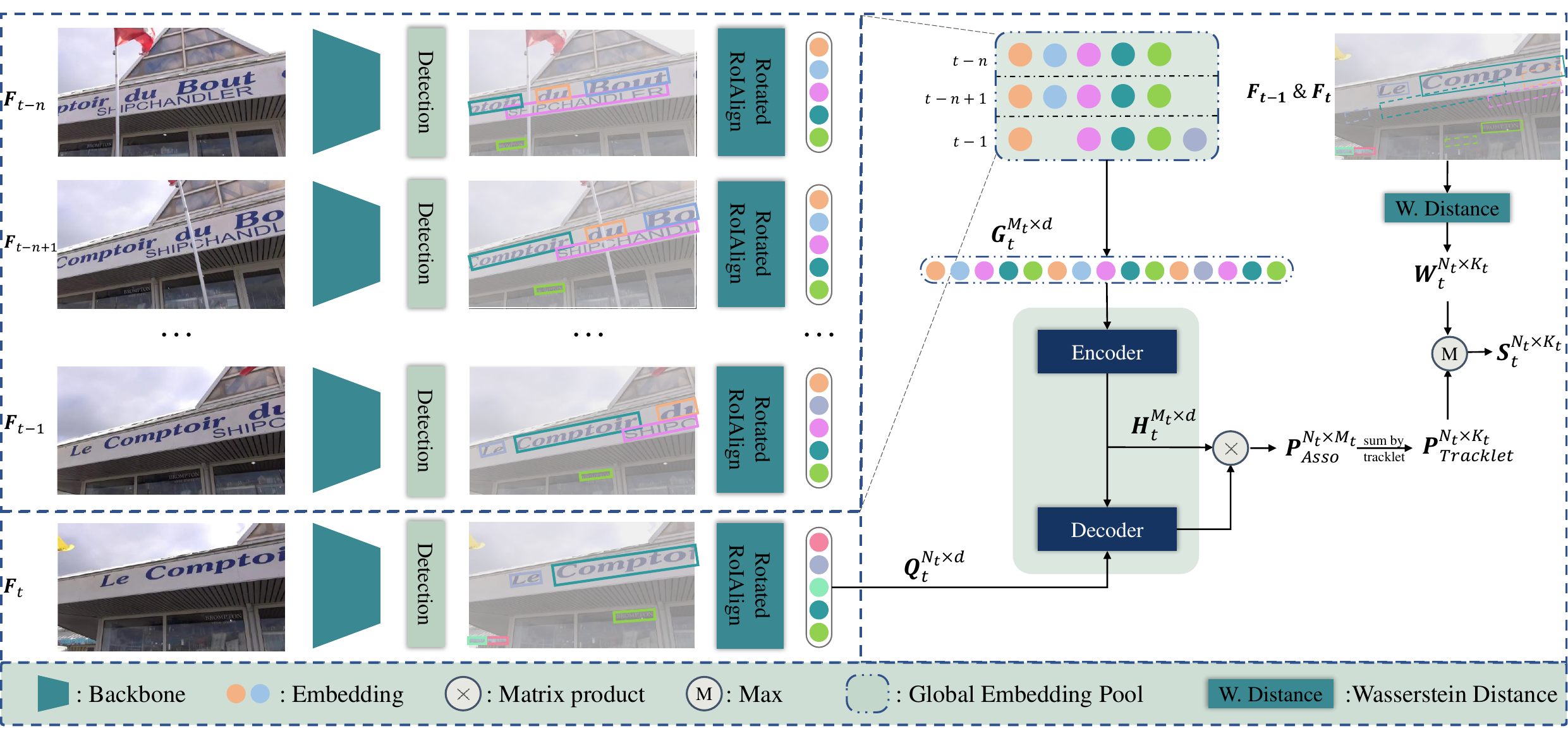}
    \caption{Overview. A global embedding pool is maintained to store historical tracking embeddings and trajectory information and is updated after each frame. With a shallow Transformer layer, we conduct associations between embeddings of the current frame and embeddings in the global embedding pool to obtain the global association score. Furthermore, a Wasserstein distance-based method is applied to measure the positional similarity between texts in frames. Some detailed architectures are ignored for clarity.}
    \label{fig:framework}
\end{figure*}
\model{} is an end-to-end framework for Video Text Spotting, which conducts global associations and adopts morphology-aware measurements. The whole framework can be seen in Fig. \ref{fig:framework}. There are three parallel heads: detection head, recognition head, and tracking head. Given a video, we first detect the potential objects as 4-point coordinates in Frame $F_t$, and then extract corresponding tracking embeddings as $\bm{{\mathit{e}}}_t^i$ for each object. We maintain a global embedding pool as $\mathbb{G}_t$, and the concatenated features of all the embeddings in $\mathbb{G}_t$ are represented as $\bm{\mathit{G}}_t \in \mathbb{R}^{M_t \times d}$, where $M_t=\sum_{i=t-L}^{t-1}N_i$ stands for the number of embeddings in $\mathbb{G}_t$. $N_t$ is the number of texts in frame $F_t$, $d$ is the dimension of each embedding, and $L$ is the sliding window size. Our tracking head calculates association scores between objects in frame $F_t$ and objects in $\mathbb{G}_t$, generating an association matrix $\bm{\mathit{P}}_{Asso} \in \mathbb{R}^{{N_t}\times{M_t}}$, and then $\bm{\mathit{P}}_{Asso}$ is turned into a tracklet-level association matrix represented as $\bm{\mathit{P_{Tracklet}}} \in \mathbb{R}^{N_t \times K_t}$, where $K_t$ is the number of tracklets in the global embedding pool $\mathbb{G}_t$. Besides, we also calculate the morphological similarities between the two adjacent frames $F_{T-1}$ and $F_T$ and output a distance matrix $\bm{\mathit{W}}_t \in \mathbb{R}^{{N_t}\times{K_{t}}}$. The two score matrices $\bm{\mathit{P_{Tracklet}}}$ and $\bm{\mathit{W}}_t$ are united by a simple max operation without any other post-processing and output the final scores depicted as $\bm{\textit{S}}_t \in \mathbb{R}^{N_t \times K_t}$ in Fig. \ref{fig:framework}. And a Hungarian algorithm is applied to assign IDs.
\begin{figure*}[!t]
    \centering
    \includegraphics[width=\linewidth]{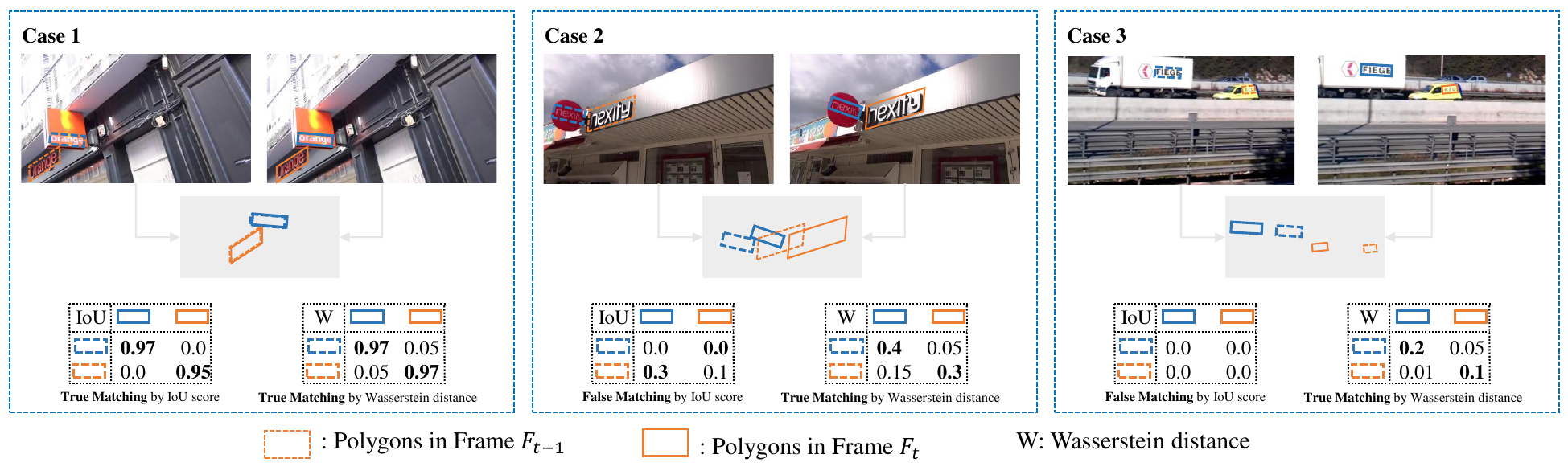}
    \caption{Three cases to demonstrate the effectiveness of Wasserstein distance. \textit{IoU}-based distance and Wasserstein distance both succeed in Case 1. But in Case 2 and Case 3, the fast movements result in the poor performance of \textit{IoU}-based distance, where Wasserstein distance produces more steady results by considering both location and morphology.}
    \label{fig:w_dis}
\end{figure*}
\subsection{Global tracking}
Motivated by GTR~\cite{GTR}, we adopt a Transformer-based network for global tracking. All the previous embeddings in the global embedding pool is encoded by a Transformer encoder as global memories. With an extraordinary long-range temporal modeling ability, Transformer is able to capture global information. Transformer decoder inputs encoded historical information as memory, with embeddings in the current frame as queries to calculate the similarity scores between the current instances and trajectories. The whole procedure can be written as:
\begin{equation}
    \bm{\mathit{H}}_t = \operatorname{Encoder}(\bm{\mathit{G}}_t),
\end{equation}
\begin{equation}
    \bm{\mathit{P}}_{Asso} = \operatorname{Decoder}(\bm{\mathit{Q}}_t, \bm{\mathit{H}}_t) \bm{\mathit{H}}_t^T, 
\end{equation}
where $\bm{\mathit{H}}_t \in \mathbb{R}^{M_t \times d}$ denotes the encoded historical temporal memory. $\bm{\mathit{P}}_{Asso} \in \mathbb{R}^{N_t \times M_t}$ is the output association matrix. $\bm{\mathit{Q}}_t \in \mathbb{R}^{N_t \times d}$ refers to the query embeddings (\textit{i.e}., embeddings of the current frame).
\\
\indent During training, we associate all the embeddings with themselves (\textit{i.e., $\bm{\mathit{Q}}_t=\bm{\mathit{G}}_t$}), generating an association matrix $\bm{\mathit{A}} \in \mathbb{R}^{{M}\times{M}}$, where $M=\sum_{t=1}^{B}N_t$ is the number of all the texts within a batch and $B$ represents the number of images in a batch, which is fixed as 16 in our experiments. For each text in each timestamp $t$, we have a vector $\bm{\mathit{a}} \in \mathbb{R}^{{N}_t + 1}$ which indicates the association scores between one query embedding $\bm{\mathit{q}}_i$ and embeddings in frame $t$. Note that the extra dimension in $\bm{\mathit{a}}$ refers to the empty association (\textit{i.e.}, the query has no matched target in this frame, usually indicating an occlusion or the end of the trajectory). A softmax function is used to transform the score into the possibility 
$\bm{\mathit{P}}_{Asso}$: \\ 
\begin{equation}
\bm{\mathit{P}}_{Asso}\left(\bm{\mathit{q}}_i, \bm{\mathit{e}}_j\right)=\frac{\exp \left( \bm{\mathit{a}}_j \right)}{\sum_{j \in\left\{\emptyset, 1, \ldots N_t\right\}} \exp \left( \bm{\mathit{a}}_j \right)}.
\end{equation}
\indent Thus, we learn the scores by minimizing the log-likelihood of each possibility. For each tracklet set $\mathbb{T}_k, k=1,2,...,K$, we assume there are $N_k$ embeddings in $\mathbb{T}_k$ and for each query embedding $\bm{\mathit{e}}_i$ in the global embedding pool $\mathbb{G}$, we calculate the loss only when $\bm{\mathit{e}}_i \in \mathbb{T}_k$. The loss writes as:
\begin{equation}
\begin{aligned}
\ell_{tracklet}\left(\mathbb{T}_k, \bm{\mathit{q}}_i \right) &= -\sum_{j=1}^{N_k} \mathbbm{1}_{\bm{\mathit{q}}_i \in \mathbb{T}_k } \cdot \log \bm{\mathit{P}}_{Asso}\left(\bm{\mathit{q}}_i, \bm{\mathit{e}}_j \right),\\
\ell_{track}&=\sum_{k=1}^{K} \sum_{i=1}^{M} \ell_{tracklet}\left(\mathbb{T}_k, \bm{\mathit{q}}_i \right), 
\end{aligned}
\end{equation}
where $\mathbbm{1}_{\bm{\mathit{q}}_i \in \mathbb{T}_k }$ is $1$ if the query embedding belongs to the tracklet set $\mathbb{T}_k$.
\\
\textbf{Semantic embeddings.} To increase the discrimination of embeddings, we also introduce semantic information to boost the performance of tracking. In detail, the embeddings fed into Transformer are extracted by Rotated RoIAlign with a shallow convolutional layer and an LSTM~\cite{LSTM} layer, followed by a fully connected layer to project features into the classes of words. The architecture writes as:\\
\begin{equation}
\begin{aligned}
    \bm{\mathit{e}}_t &= \operatorname{lstm}(\operatorname{conv}(\operatorname{r\text{-}roi}(\bm{\mathit{X}}_t))),\\
    \bm{\mathit{o}}_t &= \operatorname{fc}(\bm{\mathit{e}}_t),
\end{aligned}
\end{equation}
where $\bm{\mathit{X}}_t$ is the backbone feature map in frame $F_t$, $\bm{\mathit{e}}_t$ is the embedding fed into Transformer for associations, and $\bm{\mathit{o}}_t$ is the recognition output supervised by Connectionist Temporal Classification (CTC) \cite{CTC} loss.
\\
\textbf{Connection with previous Transformer-based methods.} Similar to previous methods \cite{TransDETR, TransVTSpotter, TransTrack, MOTR}, we also apply a Transformer architecture to carry out the association. However, there are some obvious differences: 1) \textbf{Different queries.} Queries in previous methods usually carry the identification information of trajectories, and attention is conducted between current features and track queries to output the coordinates of trajectories in the current frame. Instead, queries in \model{} carry the information of each text instance and the attention is conducted between queries and global embeddings to output an association matrix, which can gain stronger awareness of context information. 2) \textbf{Different updating strategies.} Previous works usually update the queries via an extra network (\textit{e.g.}, another Transformer) to aggregate historical information. Compared with them, our updating strategies are concise by maintaining a global embedding pool, which can be regarded as a Queue and have long-range temporal awareness. 3) \textbf{Different strategies towards new-born trajectories}. Besides track queries, previous works design extra object queries to start new-born trajectories, which may harm the performance of detection due to the competition between two kinds of queries. In comparison, we start a new track only when the association score is low, avoiding affecting detection.
\subsection{Wasserstein distances in correlation}
Concerning morphological information, we apply the Wasserstein distance to model both location similarity and shape similarity. Previous methods usually measure the location similarity between two adjacent frames by a pairwise calculation of  \textit{IoU} of each pair of bounding boxes and ignore the shape similarity, just the same as the methods in MOT. However, considering the differences from the scenes in MOT, most deformations in VTS are derived from camera motions and object motions with much fewer pose changes. In other words, the shapes of the texts are much more steady in a small time window, which can also be a strong feature for tracking. \\
\indent As demonstrated in Fig. \ref{fig:w_dis}, we use three cases to exhibit the advantage of Wasserstein distance over the \textit{IoU}. In Case 1, all texts are moving at a low speed, so the location information is enough for associations, and both \textit{IoU} and Wasserstein distance can conduct a correct match. In Case 2, the fast movement of the camera leads to the fast drift of texts. In this situation, the \textit{IoU} scores give a false positional association clue, leading to a false matching. However, the morphological differences are obvious so that the Wasserstein distance can capture the incoherence. In Case 3, the \textit{IoU} scores are both $0$ due to the fast movement of cars, while the Wasserstein distance can still perform the correct positional association. \\
\indent To obtain the awareness of both the locations and shapes, we model the polygons in different frames as Gaussian distributions and measure the similarity via distribution distance. For each predicted 4-point coordinate $b$ in two adjacent frames, we calculate pairwise Wasserstein distances between the corresponding convex hull rotated bounding boxes $\bm{\mathit{b}}(x,y,w,h,\theta)$. 
The first step is to convert the rotated box $b$ into Gaussian distribution $\mathcal{N}(\boldsymbol{\mu}, \boldsymbol{\sigma})$:
\begin{equation}
\begin{aligned}
    \boldsymbol{\mu} &= (x, y), \\
    \boldsymbol{\sigma} &=\mathbf{R S R}^{\top} \\
    &=\left(\begin{array}{cc}
    \cos \theta & -\sin \theta \\
    \sin \theta & \cos \theta
    \end{array}\right)\left(\begin{array}{cc}
    \frac{w}{2} & 0 \\
    0 & \frac{h}{2}
    \end{array}\right)\left(\begin{array}{cc}
    \cos \theta & \sin \theta \\
    -\sin \theta & \cos \theta
    \end{array}\right) \\
    &=\left(\begin{array}{cc}
    \frac{w}{2} \cos ^2 \theta+\frac{h}{2} \sin ^2 \theta & \frac{w-h}{2} \cos \theta \sin \theta \\
    \frac{w-h}{2} \cos \theta \sin \theta & \frac{w}{2} \sin ^2 \theta+\frac{h}{2} \cos ^2 \theta
    \end{array}\right),
\end{aligned}
\end{equation}
where $\mathbf{R}$ is the rotated matrix and $\mathbf{S}$ is the diagonal matrix. The Wasserstein distance between two Gaussian distributions is represented as:
\begin{equation}
\begin{aligned}
d^2 &=\left\|\boldsymbol{\mu}_1-\boldsymbol{\mu}_2\right\|_2^2+\operatorname{Tr}\left(\boldsymbol{\sigma}_1+\boldsymbol{\sigma}_2-2\left(\boldsymbol{\sigma}_1^{1 / 2} \boldsymbol{\sigma}_2 \boldsymbol{\sigma}_1^{1 / 2}\right)^{1 / 2}\right).
\end{aligned}
\end{equation}
With proper consideration of both the angles and the coordinates, Wasserstein distance can capture both location similarity and morphological similarity. Finally, to convert the distance into an applicable positional score, we have:
\begin{equation}
\begin{aligned}
\bm{\mathit{W}}\left(b_1, b_2\right) &=\bm{\mathit{W}}\left(\mathcal{N}\left(\boldsymbol{\mu}_1, \boldsymbol{\sigma}_1\right) ; \mathcal{N}\left(\boldsymbol{\mu}_2, \boldsymbol{\sigma}_2\right)\right), \\
&= 1 - \frac{\alpha d}{f\left({\boldsymbol{\sigma}_1, \boldsymbol{\sigma}_2}\right)},
\end{aligned}
\end{equation}
where $\alpha$ is a hyper-parameter, and $f$ is a function to normalize the distance. We set $f(\sigma_1, \sigma_2)=(\operatorname{Tr(\sigma_1\sigma_2)})^{1/4}$.

\subsection{Loss functions}
There are three tasks and three corresponding losses. For the detection head, we adopt L1 loss to regress the 4-point polygons and other losses are set the same as losses in YOLOX \cite{YOLOX}.
For the recognition head, we adopt Connectionist Temporal Classification (CTC) \cite{CTC} loss for texts. We also apply multi-task learning losses. The whole losses are written as:
\begin{equation}
\ell=\sigma_1 \ell_{\text {det }}+\sigma_2 \ell_{r e c}+\sigma_3 \ell_{t r a c k}+e^{-\sigma_1}+e^{-\sigma_2}+e^{-\sigma_3},
\end{equation}
where $\sigma_1, \sigma_2, \sigma_3$ are learnable parameters, and $e^{-\sigma_1}$, $e^{-\sigma_2}$, $e^{-\sigma_3}$ are regularizers for the noise terms.
\subsection{Inference}
During inference, we iteratively build the trajectories. For the initial frame $F_0$, we regard each text as the start of a trajectory. For frame $F_t$, we have a global embedding pool $\mathbb{G}_t$ which stores all the previous embeddings in a sliding window, and the corresponding embedding matrix writes as $\bm{\mathit{G}}_t$ (\textit{i.e.,} the concatenated embeddings). Assume there are $K_t$ trajectories in $\mathbb{G}_t$, and each trajectory has $N_k$ embeddings. We regard embeddings in the current frame $F_t$ as $\bm{\mathit{Q}}_t$ and the corresponding embedding matrix as $\bm{\mathit{Q}}_t$. The association score $A_t$ is calculated between $\bm{\mathit{Q}}_t$ and $\bm{\mathit{G}}_t$, and converted to a possibility matrix $\bm{\mathit{P}}_{Asso}$ by a softmax function.
Then a tracklet-wise sum is applied to calculate the possibility of each tracklet. For each $\bm{\mathit{q}}_i \in \mathbb{Q}_t$ we have:
\begin{equation}
\bm{\mathit{P}}_{Tracklet}(\bm{\mathit{q}}_i, \bm{\mathit{G}}_t) = \sum_{j=1}^{N_k} \bm{\mathit{P}}_{Asso}(\bm{\mathit{q}}_i, \bm{\mathit{e}}_j),
\end{equation}
where $\mathbf{e}_j$ is the embedding in each trajectory. After calculating all the embeddings in $\mathbb{Q}_t$, we can finally obtain a matrix $\bm{\mathit{P}}_{Tracklet} \in \mathbb{R}^{N_t \times K_t}$, which refers to the possibility that the embedding belongs to the tracklet. Also, Wasserstein distance between frame $F_{t-1}$ and $F_t$ is calculated as a positional and morphological similarity score $\bm{\mathit{W}}_t$, and the final output is $\mathrm{max}(\bm{\mathit{P}}_{Tracklet}, \bm{\mathit{W}}_t)$. A Hungarian algorithm is applied to ensure the ID assignment is unique for each text. 
\section{Experiments}

\subsection{Implemented details}
We adopt ResNet-50 \cite{ResNet} with FPN \cite{FPN} layers as our backbone and use the checkpoint pretrained on ImageNet~\cite{imagenet1, imagenet2}. The architecture of the detection head is borrowed from YOLOX~\cite{YOLOX} with an extra branch to regress the polygons. The tracking head is a lightweight architecture with only a one-layer Transformer. All experiments are conducted on Tesla V100 GPUs. We first pretrain the model on COCOText~\cite{cocotext} and apply Random Crop, Random Resize, Random Color Jittering, and Pseudo Track~\cite{CenterTrack} for data augmentation. Then it is fine-tuned on other datasets. The batch size is fixed as 16 when training and random sampling within a clip is applied to make sure images in a batch are from the same video clip. During inference, we resize the images with the shorter side fixed and the ratio of images kept. 
\begin{table}[!t]
    \begin{center}
    \setlength{\tabcolsep}{1pt}
    \resizebox{\linewidth}{!}{ 
    \begin{tabular}{llccccc}
    \toprule
        Dataset     & Methods   & MOTA$\uparrow$    & MOTP$\uparrow$    & IDF1$\uparrow$    & MM$\uparrow$  & ML$\downarrow$ \\
        \midrule
        \multirow{7}{5em}{ICDAR2015 video} 
        & AJOU~\cite{AJOU}                      & 16.4  & 72.7  & 36.1  & 14.1  & 62.0\\
        & Free~\cite{Free}                      & 43.2  & 76.7  & 57.9  & 36.6  & 44.4\\
        & SAVTD~\cite{SAVTD}                    & 44.1  & 75.2  & 58.2  & 44.8  & 29.0\\
        & SVRep~\cite{SVRep}                    & 49.5  & 73.9  & 66.1  & 44.9  & 27.1\\
        & CoText~\cite{CoText}                  & 51.4  & 73.6  & 68.6  & 49.6  & \textbf{23.5}\\
        & TransVTSpotter~\cite{TransVTSpotter}  & 44.1 & 75.8 & 57.3 & 34.3 & 33.7\\
        & TransDETR~\cite{TransDETR}            & 47.7 & 74.1 & 65.5 & 42.0 & 32.1\\
        & Ours                                  & \textbf{56.0} & \textbf{77.4} & \textbf{70.5} & \textbf{49.7} & 27.3\\
        \midrule
        \multirow{5}{5em}{ICDAR2013 video} & YORO~\cite{YORO} & 47.3 & 73.7 & 62.5 & 33.1 & 45.3\\
        & SVRep~\cite{SVRep} & 53.2 & 76.7 & 65.1 & 38.2 & 33.2\\
        & CoText~\cite{CoText} & 55.8 & 76.4 & 68.1 & 44.6 & 28.7\\
        & TransDETR~\cite{TransDETR} & 54.7 & 76.6 & 67.2 & 43.5 & 33.2\\
        & Ours & \textbf{56.3} & \textbf{78.7} & \textbf{68.6} & \textbf{46.0} & \textbf{28.6}\\
        \midrule
        \multirow{6}{5em}{Minetto} & SAVTD~\cite{SAVTD} & 83.5 & 76.8 & - & - & -\\
        & SVRep~\cite{SVRep} & 86.3 & \textbf{81.0}  & 83.9 & \textbf{96.4} & \textbf{0}\\
        & CoText~\cite{CoText} & 86.9 & 80.6  & 83.9 & 87.7 & 0\\
        & TransVTSpotter~\cite{TransVTSpotter} & 84.1 & 77.6 & 74.7  & - & -\\
        & TransDETR~\cite{TransDETR} & 84.1 & 57.9 & 76.7  & 36.6 & 44.4\\
        & Ours & \textbf{87.1} & 80.6 & \textbf{84.2} & 89.3 & 3.6\\
    \bottomrule
    \end{tabular}
        
    }
    \end{center}
    \caption{Video Text Tracking on different datasets. Our proposed method outperforms previous methods by a large margin.}
    \label{r50}
\end{table}
\begin{table}[!ht]
    \begin{center}
    \setlength{\tabcolsep}{1pt}
    \begin{tabular}{lccccc}
    \toprule
        Methods & MOTA$\uparrow$ & MOTP$\uparrow$ & IDF1$\uparrow$ & MM$\uparrow$ & ML$\downarrow$\\
        \midrule
        Free~\cite{Free} & 53.0 & 74.9 & 61.9 & 45.5 & 35.9\\
        CoText~\cite{CoText} & 59.0 & 74.5 & 72.0 & 48.6 & 26.4\\
        TransVTSpotter~\cite{TransVTSpotter} & 53.2 & 74.9 & 61.5 & - & -\\
        TransDETR~\cite{TransDETR} & 58.4 & 75.2 & 70.4 & 32.0 & \textbf{20.8}\\
        TransDETR (aug)~\cite{TransDETR} & 60.9 & 74.6 & 72.8 & 33.6 & 20.8\\
        OURS & \textbf{62.5} & \textbf{78.2} & \textbf{74.2} & \textbf{51.0} & 22.0\\
    \bottomrule
    \end{tabular}
    \end{center}
    \caption{Video Text Spotting on ICDAR2015 video dataset.  Our \model{} also achieves leading performance.}
    \label{vts}
\end{table}
\begin{table}[!t]
    \begin{center}
    \begin{tabular}{lccc}
    \toprule
        Methods & Precision$\uparrow$ & Recall$\uparrow$ & F-score$\uparrow$\\
        \midrule
        Free~\cite{Free} & 79.7 & 68.4 & 73.6\\
        SVRep~\cite{SVRep} & 81.2 & 68.3 & 74.2 \\
        TransDETR~\cite{TransDETR} & 80.6 & \textbf{70.2} & 75.0\\
        Ours & \textbf{89.8} & 64.5& \textbf{75.1}\\
    \bottomrule
    \end{tabular}
    \end{center}
    \caption{Detection performance on ICDAR13 video dataset.}   
    \label{fscore}
\end{table}
\subsection{Datasets and metrics}
Following previous protocols, we evaluate our methods on several different datasets.\\
\textbf{COCOTEXT}. COCOText has 60000+ images and more than 200000 texts, which is usually used as a pretraining dataset in VTS. \\
\textbf{ICDAR2015 video and ICDAR2013 video}. ICDAR2015 video contains 25 clips for training and 24 clips for testing. Most scenes are street views with tens of texts in one image. ICDAR2013 video is a sub-dataset of ICDAR2015 video. \\
\textbf{Minetto}. Minetto is a small dataset that contains 5 videos harvested outdoors. Without a training split, it is used as a test dataset in previous methods.\\ 
\textbf{Metrics}. Following previous protocols~\cite{MOTMetrics}, we adopt the metrics inherited from MOT. Different from MOT, metrics in Video Text Tracking adopt the \textit{IoU} between polygons to measure the similarity of two instances. Three metrics, MOTA, MOTP, and IDF1, are mainly used to evaluate performance. MOTA measures a comprehensive performance of both the detection and tracking performance, and MOTP mainly concerns the ability to fit the bounding boxes. IDF1 only measures the ability of tracking. Besides, we also adopt Mostly-Matched (MM) and Mostly-Lost (ML) to evaluate the completeness of trajectories. In Video Text Spotting, we also use the same metrics to measure the performance, but the similarity between instances is calculated by the edit distance between texts.
\subsection{State-of-the-art comparisons}
To verify the effectiveness of the proposed \model{}, we compare its performance with several SOTA methods. SOTA methods can be divided into the following three categories.\\
1) Local association based. Previous works \cite{SVRep, CoText, Free, AJOU} only conduct local associations by matching texts in two adjacent frames.\\
2) Video Object Detection based. \cite{SAVTD, Free} apply VOD based methods to encode temporal information into features.\\
3) Transformer based. \cite{TransDETR, TransVTSpotter} apply queries to decode the corresponding results in each frame with a query updating scheme to obtain temporal awareness.
\\
\textbf{Video Text Tracking}. Video Text Tracking is a core task to measure the performance of all methods. Thus, we evaluate our method in VTT and compare with other methods on several public datasets. On the most commonly used dataset ICDAR2015 video, we outperform previous works by a large margin. We obtain 4.6 absolute improvement over the previous state-of-the-art method and 8.3 absolute improvement over the previous Transformer-based method, which proves the effectiveness of the proposed \model{}. Besides, our method also achieves leading performance on MOTP and IDF1. On ICDAR2013 video, our \model{} achieves top performance on all metrics. On a smaller dataset Minetto, we also achieve SOTA results.
\\
\textbf{Video Text Spotting}. 
Video Text Spotting concerns the tracking performance and the recognition results. As shown in Tab. \ref{vts}, compared with the previous SOTA method TransDETR (aug),~\cite{TransDETR} our \model{} obtains 1.6 absolute improvement on MOTA, 3.6 absolute improvement on MOTP, and 1.4 absolute improvement on IDF1. Therefore, with a simple and shallow recognition head, our \model{} can also achieve great performance. 
\\
\textbf{Text Detection}. 
We also evaluate the detection performances of methods in Tab. \ref{fscore}. The detection performance of \model{} is comparable with the SOTA methods.
\subsection{Ablation studies}
\begin{table}[!t]
    \begin{center}
    \setlength{\tabcolsep}{2pt}
    \resizebox{\linewidth}{!}{
    \begin{tabular}{cccccccc}
    \toprule
        Window size & MOTA$\uparrow$ & MOTP$\uparrow$ & IDF1$\uparrow$ & MM$\uparrow$ & ML$\downarrow$ & Asso. T (ms)$\downarrow$\\
        \midrule
        2 & 55.1 & \textbf{77.5} & 63.0 & 46.8 & 30.1 & \textbf{3.1}\\
        4 & 55.9 & 77.4 & 68.3 & 49.4 & 28.2 & 3.2\\
        8 & \textbf{56.0} & 77.4 & 70.5 & \textbf{49.7} & 27.3 & 3.6\\
        16 & 55.7 & 77.4 & \textbf{71.0} & 49.1 & \textbf{27.1} & 3.9\\
    \bottomrule
    \end{tabular}
    }
    \end{center}
   
    \caption{Ablation study on the sliding window size. As the window size goes larger, the tracking performance tends to improve.}
     \label{window}
    
\end{table}

   

\begin{table}[!t]
    \begin{center}
    \setlength{\tabcolsep}{2pt}
    \begin{tabular}{lccccc}
    \toprule
        Methods & MOTA$\uparrow$ & MOTP$\uparrow$ & IDF1$\uparrow$ & MM$\uparrow$ & ML$\downarrow$\\
        \midrule
        \textit{w/o.} distance & 55.5 & 77.4 & 70.0 & 47.3 & 28.1\\
        \textit{w.} IoU & 55.8& 77.4 & \textbf{70.6} & 49.0 & \textbf{27.1}\\
        \textit{w.} L1 & 55.5 & \textbf{77.5} & 70.3 & 48.3 & 27.9\\
        \textit{w.} L2 & 55.8 & 77.4 & 70.2 & 49.7 & 27.6\\
        \textit{w.} Wasserstein & \textbf{56.0} & 77.4 & 70.5 & \textbf{49.7} & 27.3\\
        \midrule   
        \textit{w.} self-attn & 53.2 & 77.1 & 69.8 & \textbf{52.8} & \textbf{25.4}\\
        \textit{w/o.} self-attn & \textbf{56.0} & \textbf{77.4} & \textbf{70.5} & 49.7 & 27.3\\
        \midrule
        \textit{w.} $\bm{\mathit{P}}_{Tracklet}$ & 55.5 & 77.4 & 70.0 & 47.3 & 28.1\\
        \textit{w.} $\bm{\mathit{W}}_t$ & 43.2 & \textbf{77.8} & 50.2 & 33.0 & 40.2\\
        \textit{w.} $max(\bm{\mathit{P}}, \bm{\mathit{W}})$ & \textbf{56.0} & 77.4 & \textbf{70.5} & \textbf{49.7} & \textbf{27.3}\\
        \midrule
        \textit{w/o.} semantic & 53.9 & 77.4 & 66.9 & 47.7 & 28.0\\
        \textit{w.} semantic &\textbf{56.0} & \textbf{77.4} & \textbf{70.5} & \textbf{49.7} & \textbf{27.3}\\
        
    \bottomrule
    \end{tabular}
    \end{center}
    
    \caption{Experiments on positional distance (\textbf{Row 1-5}), self attention layer in decoder (\textbf{Row 6-7}), max operation (\textbf{Row 8-10}), and semantic embeddings (\textbf{Row 11-12}).}
    \label{rebuttalablation}
    
\end{table}

To verify the effectiveness of the components of \model{}, we conduct several ablation studies on ICDAR2015 video, as shown in Tab. \ref{window}-\ref{rebuttalablation}.\\
\textbf{Sliding window size}. We study how the size of the sliding window impacts the final results during the inference stage. As shown in Tab. \ref{window}, we can see a trend that the tracking performance goes up when the window length increases. 
When the window length is $2$,  the situation only involves associations between the two adjacent frames (\textit{i.e.,} local associations), and the performance is much worse than when using a larger window size.
Note that the metric MOTA concerns both the detection performance and tracking performance, and IDF1 only focuses on the tracking performance. Besides the temporal awareness obtained in training process, a longer window during inference stage also brings longer range of temporal information, leading to a sharp increase in IDF1, which proves the contribution of global embeddings to the final tracking performance.
Therefore, we adopt 8 as the default sliding window size.\\
\textbf{Distance measurement.} We adopt Wasserstein distance for the measurement of positional association instead of \textit{IoU}. As mentioned above, Wasserstein distance pays attention to both the location and morphological features so that it outperforms \textit{IoU} when applied as a positional distance measurement in VTS. As shown in Tab. \ref{rebuttalablation} (Row 1-5), positional distance can improve the performance, and with Wasserstein distance, the proposed \model{} can achieve better results than with \textit{IoU} or other common distances (\textit{e.g.}, L1 or L2 distance).\\
\textbf{Attention layer.} The encoder layer is one multi-head attention layer and the decoder layer includes one cross-attention layer without a self-attention layer. As shown in Tab. \ref{rebuttalablation} (Row 4-5), we do not observe obvious improvement of the overall performance when introducing self-attention layer and thus we remove it.
\\
 \textbf{Max operation.} We show extra experiments in Tab. \ref{rebuttalablation} (Row 8-10). Both $\bm{\mathit{P}}_{Tracklet}$ and $\bm{\mathit{W}}_t$ contribute to the performance.
 \\
 \textbf{Semantic embeddings}. We use the embeddings before the final fully-connected layer for associations. The embeddings we select carry semantic information and can provide some prior information and boost tracking performance. To verify the effectiveness, we only feed Transformer the features after Rotated RoIAlign to explore the influence of semantic information. As shown in Tab. \ref{rebuttalablation} (Row 11-12), the ablation studies show the effectiveness of the proposed \model{}. \\
\textbf{Speed analysis.} Our \model{} runs at around $20$ FPS on a single Tesla V100 GPU. The backbone, FPN layers, detection head, and recognition head take about $48$ ms per frame, and the tracking procedure takes about $3.6$ ms per frame. By maintaining a global embedding pool, we do not have to repeatedly extract the embeddings from each frame. As shown in Tab. \ref{window}, Asso.~T refers to the time cost by the tracking procedure (\textit{i.e.}, the inference of Transformer and associations). When the window size goes larger, we only observe a slight increment in the time consumed. \\
\subsection{Visualization}
\begin{figure*}[!ht]
    \centering
    \includegraphics[width=\textwidth]{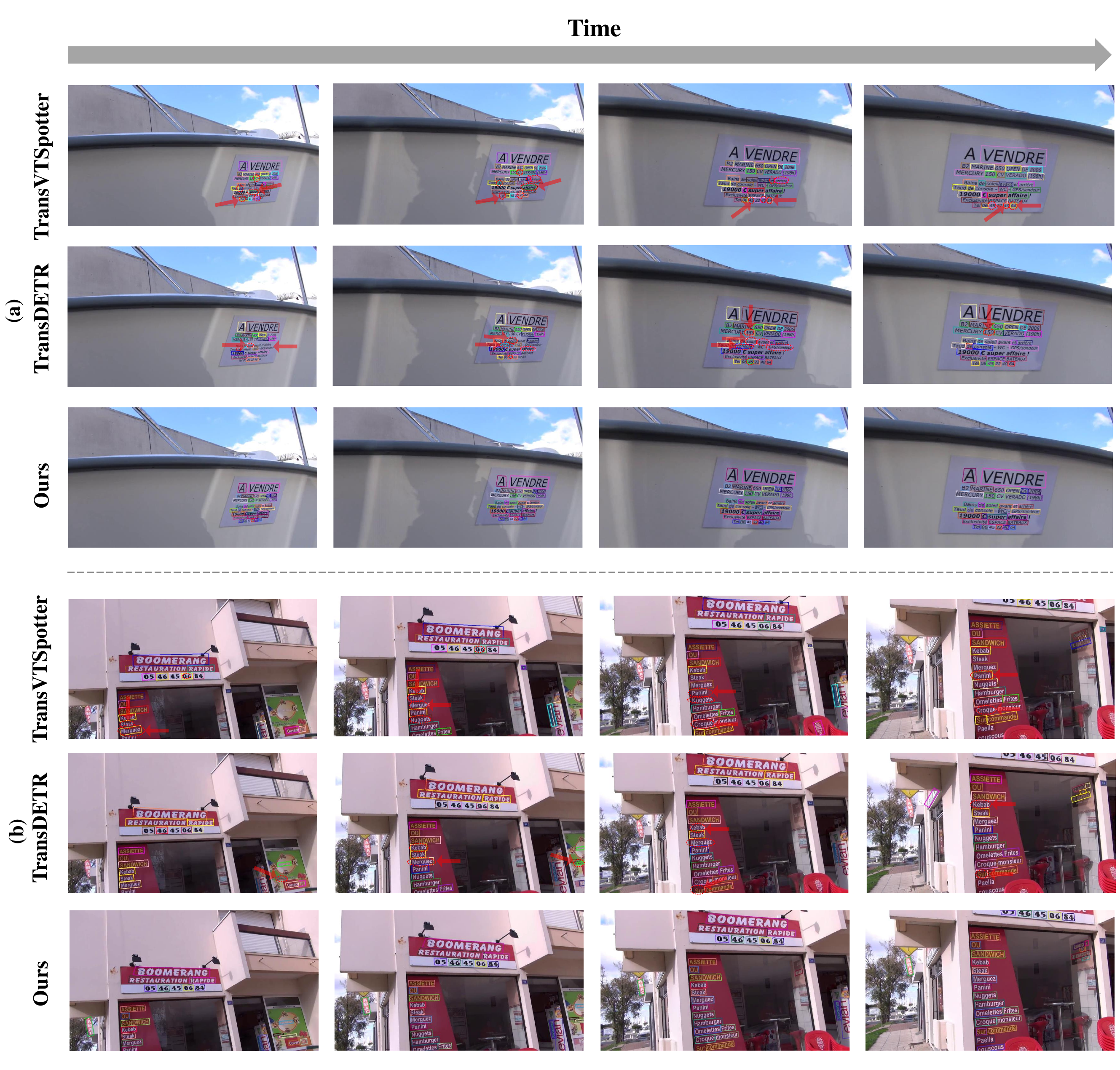}
    \caption{We demonstrate the results of previous Transformer-based methods~\cite{TransVTSpotter, TransDETR} and our \model{}. Different IDs are represented in different colors. Some of the false results (\textit{e.g.,} FNs, ID switches, and IDFs) are marked with a dotted red circle and pointed out by a red arrow. Apparently, our \model{} performs better than previous Transformer-based methods especially in crowded scenes.}
    \label{fig:results}
\end{figure*}

As demonstrated in Fig. \ref{fig:results}, we select two videos to show the advantage of our \model{}. From an overall perspective, our \model{} performs fewer false assignments in tracking and fewer FNs in detection. As presented in video (a), when facing crowded scenes with motion blur, TransVTSpotter and TransDETR fail to perform right assignments due to excessive interfering texts. We could notice that the polygon colors of the texts (\textit{e.g.,} ``CV", ``de", ``super", ``Tel") are changed through time, indicating many ID switches. On the contrary, our \model{} has a much more steady performance without any ID switch. As presented in video (b), the architecture of TransDETR and TransVTSpotter have some limitations in detection, resulting in False Negatives in some frames. Without a global view, TransVTSpotter and TransDETR tend to conduct incorrect assignments especially when the fractures of trajectories take place. Our detection performance is relatively independent from tracking, and our \model{} has a global view on historical information, leading to better performance on both detection and tracking.
\section{Limitations and conclusion}
\noindent \textbf{Limitations. }Though \model{} achieves great performance on VTS, we find the detector fails in detecting texts when facing severe motion blur and extreme sizes, which leads to fractures of trajectories. Our detector is less robust towards deterioration compared with our tracking method. We would consider how to apply trajectory information to improve detection performance in our future work.\\
\noindent \textbf{Conclusion. }In this work, we propose a Transformer-based global text spotting method \model{}. We explore how to fully exploit temporal information and morphological information with a global association module and Wasserstein distance, respectively. We also conduct
extensive experiments on the public datasets to verify the effectiveness of our method. We hope our work can shed light on future research on the utilization of temporal information and morphological information in Video Text Spotting.

{\small
\bibliographystyle{ieee_fullname}

\begin{thebibliography}{10}\itemsep=-1pt

\bibitem{SORT}
Alex Bewley, Zongyuan Ge, Lionel Ott, Fabio Ramos, and Ben Upcroft.
\newblock Simple online and realtime tracking.
\newblock In {\em 2016 IEEE international conference on image processing
  (ICIP)}, pages 3464--3468. IEEE, 2016.

\bibitem{MEGA}
Yihong Chen, Yue Cao, Han Hu, and Liwei Wang.
\newblock Memory enhanced global-local aggregation for video object detection.
\newblock In {\em Proceedings of the IEEE/CVF conference on computer vision and
  pattern recognition}, pages 10337--10346, 2020.

\bibitem{YORO}
Zhanzhan Cheng, Jing Lu, Yi Niu, Shiliang Pu, Fei Wu, and Shuigeng Zhou.
\newblock You only recognize once: Towards fast video text spotting.
\newblock In {\em Proceedings of the 27th ACM International Conference on
  Multimedia}, pages 855--863, 2019.

\bibitem{Free}
Zhanzhan Cheng, Jing Lu, Baorui Zou, Liang Qiao, Yunlu Xu, Shiliang Pu, Yi Niu,
  Fei Wu, and Shuigeng Zhou.
\newblock Free: A fast and robust end-to-end video text spotter.
\newblock {\em IEEE Transactions on Image Processing}, 30:822--837, 2020.

\bibitem{imagenet1}
Jia Deng, Wei Dong, Richard Socher, Li-Jia Li, Kai Li, and Li Fei-Fei.
\newblock Imagenet: A large-scale hierarchical image database.
\newblock In {\em 2009 IEEE conference on computer vision and pattern
  recognition}, pages 248--255. Ieee, 2009.

\bibitem{RDN}
Jiajun Deng, Yingwei Pan, Ting Yao, Wengang Zhou, Houqiang Li, and Tao Mei.
\newblock Relation distillation networks for video object detection.
\newblock In {\em ICCV}, pages 7023--7032, 2019.

\bibitem{SAVTD}
Wei Feng, Fei Yin, Xu-Yao Zhang, and Cheng-Lin Liu.
\newblock Semantic-aware video text detection.
\newblock In {\em Proceedings of the IEEE/CVF Conference on Computer Vision and
  Pattern Recognition}, pages 1695--1705, 2021.

\bibitem{STCM}
Yuzhe Gao, Xing Li, Jiajian Zhang, Yu Zhou, Dian Jin, Jing Wang, Shenggao Zhu,
  and Xiang Bai.
\newblock Video text tracking with a spatio-temporal complementary model.
\newblock {\em IEEE Transactions on Image Processing}, 30:9321--9331, 2021.

\bibitem{YOLOX}
Zheng Ge, Songtao Liu, Feng Wang, Zeming Li, and Jian Sun.
\newblock Yolox: Exceeding yolo series in 2021.
\newblock {\em arXiv preprint arXiv:2107.08430}, 2021.

\bibitem{fastrcnn}
Ross Girshick.
\newblock Fast r-cnn.
\newblock In {\em Proceedings of the IEEE international conference on computer
  vision}, pages 1440--1448, 2015.

\bibitem{CTC}
Alex Graves, Santiago Fern{\'a}ndez, Faustino Gomez, and J{\"u}rgen
  Schmidhuber.
\newblock Connectionist temporal classification: labelling unsegmented sequence
  data with recurrent neural networks.
\newblock In {\em Proceedings of the 23rd international conference on Machine
  learning}, pages 369--376, 2006.

\bibitem{maskrcnn}
Kaiming He, Georgia Gkioxari, Piotr Doll{\'a}r, and Ross Girshick.
\newblock Mask r-cnn.
\newblock In {\em Proceedings of the IEEE international conference on computer
  vision}, pages 2961--2969, 2017.

\bibitem{ResNet}
Kaiming He, Xiangyu Zhang, Shaoqing Ren, and Jian Sun.
\newblock Deep residual learning for image recognition.
\newblock In {\em Proceedings of the IEEE conference on computer vision and
  pattern recognition}, pages 770--778, 2016.

\bibitem{LSTM}
Sepp Hochreiter and J{\"u}rgen Schmidhuber.
\newblock Long short-term memory.
\newblock {\em Neural computation}, 9(8):1735--1780, 1997.

\bibitem{ICDAR2015}
Dimosthenis Karatzas, Lluis Gomez-Bigorda, Anguelos Nicolaou, Suman Ghosh,
  Andrew Bagdanov, Masakazu Iwamura, Jiri Matas, Lukas Neumann,
  Vijay~Ramaseshan Chandrasekhar, Shijian Lu, et~al.
\newblock Icdar 2015 competition on robust reading.
\newblock In {\em 2015 13th international conference on document analysis and
  recognition (ICDAR)}, pages 1156--1160. IEEE, 2015.

\bibitem{ICDAR2013}
Dimosthenis Karatzas, Faisal Shafait, Seiichi Uchida, Masakazu Iwamura,
  Lluis~Gomez i Bigorda, Sergi~Robles Mestre, Joan Mas, David~Fernandez Mota,
  Jon~Almazan Almazan, and Lluis~Pere De~Las~Heras.
\newblock Icdar 2013 robust reading competition.
\newblock In {\em 2013 12th international conference on document analysis and
  recognition}, pages 1484--1493. IEEE, 2013.

\bibitem{AJOU}
Hyung~Il Koo and Duck~Hoon Kim.
\newblock Scene text detection via connected component clustering and nontext
  filtering.
\newblock {\em IEEE transactions on image processing}, 22(6):2296--2305, 2013.

\bibitem{metriclearning}
Brian Kulis et~al.
\newblock Metric learning: A survey.
\newblock {\em Foundations and Trends{\textregistered} in Machine Learning},
  5(4):287--364, 2013.

\bibitem{SVRep}
Zhuang Li, Weijia Wu, Mike~Zheng Shou, Jiahong Li, Size Li, Zhongyuan Wang, and
  Hong Zhou.
\newblock Contrastive learning of semantic and visual representations for text
  tracking.
\newblock {\em arXiv preprint arXiv:2112.14976}, 2021.

\bibitem{FPN}
Tsung-Yi Lin, Piotr Doll{\'a}r, Ross Girshick, Kaiming He, Bharath Hariharan,
  and Serge Belongie.
\newblock Feature pyramid networks for object detection.
\newblock In {\em Proceedings of the IEEE conference on computer vision and
  pattern recognition}, pages 2117--2125, 2017.

\bibitem{FOTS}
Xuebo Liu, Ding Liang, Shi Yan, Dagui Chen, Yu Qiao, and Junjie Yan.
\newblock Fots: Fast oriented text spotting with a unified network.
\newblock In {\em Proceedings of the IEEE conference on computer vision and
  pattern recognition}, pages 5676--5685, 2018.

\bibitem{FCN}
Jonathan Long, Evan Shelhamer, and Trevor Darrell.
\newblock Fully convolutional networks for semantic segmentation.
\newblock In {\em Proceedings of the IEEE conference on computer vision and
  pattern recognition}, pages 3431--3440, 2015.

\bibitem{Minetto}
Rodrigo Minetto, Nicolas Thome, Matthieu Cord, Neucimar~J Leite, and Jorge
  Stolfi.
\newblock Snoopertrack: Text detection and tracking for outdoor videos.
\newblock In {\em 2011 18th IEEE international conference on image processing},
  pages 505--508. IEEE, 2011.

\bibitem{MOTMetrics}
Ergys Ristani, Francesco Solera, Roger Zou, Rita Cucchiara, and Carlo Tomasi.
\newblock Performance measures and a data set for multi-target, multi-camera
  tracking.
\newblock In {\em European conference on computer vision}, pages 17--35.
  Springer, 2016.

\bibitem{imagenet2}
Olga Russakovsky, Jia Deng, Hao Su, Jonathan Krause, Sanjeev Satheesh, Sean Ma,
  Zhiheng Huang, Andrej Karpathy, Aditya Khosla, Michael Bernstein, et~al.
\newblock Imagenet large scale visual recognition challenge.
\newblock {\em International journal of computer vision}, 115(3):211--252,
  2015.

\bibitem{LongRange}
Mykhailo Shvets, Wei Liu, and Alexander~C Berg.
\newblock Leveraging long-range temporal relationships between proposals for
  video object detection.
\newblock In {\em ICCV}, pages 9756--9764, 2019.

\bibitem{TransTrack}
Peize Sun, Jinkun Cao, Yi Jiang, Rufeng Zhang, Enze Xie, Zehuan Yuan, Changhu
  Wang, and Ping Luo.
\newblock Transtrack: Multiple object tracking with transformer.
\newblock {\em arXiv preprint arXiv:2012.15460}, 2020.

\bibitem{CTPN}
Zhi Tian, Weilin Huang, Tong He, Pan He, and Yu Qiao.
\newblock Detecting text in natural image with connectionist text proposal
  network.
\newblock In {\em European conference on computer vision}, pages 56--72.
  Springer, 2016.

\bibitem{cocotext}
Andreas Veit, Tomas Matera, Lukas Neumann, Jiri Matas, and Serge Belongie.
\newblock Coco-text: Dataset and benchmark for text detection and recognition
  in natural images.
\newblock {\em arXiv preprint arXiv:1601.07140}, 2016.

\bibitem{PTSEFormer}
Han Wang, Jun Tang, Xiaodong Liu, Shanyan Guan, Rong Xie, and Li Song.
\newblock Ptseformer: Progressive temporal-spatial enhanced transformer towards
  video object detection.
\newblock {\em arXiv preprint arXiv:2209.02242}, 2022.

\bibitem{PSENet}
Wenhai Wang, Enze Xie, Xiang Li, Wenbo Hou, Tong Lu, Gang Yu, and Shuai Shao.
\newblock Shape robust text detection with progressive scale expansion network.
\newblock In {\em Proceedings of the IEEE/CVF Conference on Computer Vision and
  Pattern Recognition}, pages 9336--9345, 2019.

\bibitem{Panpp}
Wenhai Wang, Enze Xie, Xiang Li, Xuebo Liu, Ding Liang, Zhibo Yang, Tong Lu,
  and Chunhua Shen.
\newblock Pan++: Towards efficient and accurate end-to-end spotting of
  arbitrarily-shaped text.
\newblock {\em IEEE Transactions on Pattern Analysis and Machine Intelligence},
  44(9):5349--5367, 2021.

\bibitem{Pan}
Wenhai Wang, Enze Xie, Xiaoge Song, Yuhang Zang, Wenjia Wang, Tong Lu, Gang Yu,
  and Chunhua Shen.
\newblock Efficient and accurate arbitrary-shaped text detection with pixel
  aggregation network.
\newblock In {\em Proceedings of the IEEE/CVF International Conference on
  Computer Vision}, pages 8440--8449, 2019.

\bibitem{JDE}
Zhongdao Wang, Liang Zheng, Yixuan Liu, Yali Li, and Shengjin Wang.
\newblock Towards real-time multi-object tracking.
\newblock In {\em European Conference on Computer Vision}, pages 107--122.
  Springer, 2020.

\bibitem{DeepSORT}
Nicolai Wojke, Alex Bewley, and Dietrich Paulus.
\newblock Simple online and realtime tracking with a deep association metric.
\newblock In {\em 2017 IEEE international conference on image processing
  (ICIP)}, pages 3645--3649. IEEE, 2017.

\bibitem{SELSA}
Haiping Wu, Yuntao Chen, Naiyan Wang, and Zhaoxiang Zhang.
\newblock Sequence level semantics aggregation for video object detection.
\newblock In {\em ICCV}, pages 9217--9225, 2019.

\bibitem{TransVTSpotter}
Weijia Wu, Yuanqiang Cai, Debing Zhang, Sibo Wang, Zhuang Li, Jiahong Li, Yejun
  Tang, and Hong Zhou.
\newblock A bilingual, openworld video text dataset and end-to-end video text
  spotter with transformer.
\newblock {\em arXiv preprint arXiv:2112.04888}, 2021.

\bibitem{CoText}
Wejia Wu, Zhuang Li, Jiahong Li, Chunhua Shen, Hong Zhou, Size Li, Zhongyuan
  Wang, and Ping Luo.
\newblock Real-time end-to-end video text spotter with contrastive
  representation learning.
\newblock {\em arXiv preprint arXiv:2207.08417}, 2022.

\bibitem{TransDETR}
Weijia Wu, Debing Zhang, Ying Fu, Chunhua Shen, Hong Zhou, Yuanqiang Cai, and
  Ping Luo.
\newblock End-to-end video text spotting with transformer.
\newblock {\em arXiv preprint arXiv:2203.10539}, 2022.

\bibitem{GWD}
Xue Yang, Junchi Yan, Qi Ming, Wentao Wang, Xiaopeng Zhang, and Qi Tian.
\newblock Rethinking rotated object detection with gaussian wasserstein
  distance loss.
\newblock In {\em International Conference on Machine Learning}, pages
  11830--11841. PMLR, 2021.

\bibitem{online}
Hongyuan Yu, Yan Huang, Lihong Pi, Chengquan Zhang, Xuan Li, and Liang Wang.
\newblock End-to-end video text detection with online tracking.
\newblock {\em Pattern Recognition}, 113:107791, 2021.

\bibitem{MOTR}
Fangao Zeng, Bin Dong, Tiancai Wang, Xiangyu Zhang, and Yichen Wei.
\newblock Motr: End-to-end multiple-object tracking with transformer.
\newblock {\em arXiv preprint arXiv:2105.03247}, 2021.

\bibitem{FairMOT}
Yifu Zhang, Chunyu Wang, Xinggang Wang, Wenjun Zeng, and Wenyu Liu.
\newblock Fairmot: On the fairness of detection and re-identification in
  multiple object tracking.
\newblock {\em International Journal of Computer Vision}, 129(11):3069--3087,
  2021.

\bibitem{Deeptext}
Z Zhong, L Jin, S Zhang, and Z Feng.
\newblock Deeptext: A unified framework for text proposal generation and text
  detection in natural images. arxiv 2016.
\newblock {\em arXiv preprint arXiv:1605.07314}.

\bibitem{TransVOD}
Qianyu Zhou, Xiangtai Li, Lu He, Yibo Yang, Guangliang Cheng, Yunhai Tong,
  Lizhuang Ma, and Dacheng Tao.
\newblock Transvod: End-to-end video object detection with spatial-temporal
  transformers.
\newblock {\em arXiv preprint arXiv:2201.05047}, 2022.

\bibitem{CenterTrack}
Xingyi Zhou, Vladlen Koltun, and Philipp Kr{\"a}henb{\"u}hl.
\newblock Tracking objects as points.
\newblock In {\em European Conference on Computer Vision}, pages 474--490.
  Springer, 2020.

\bibitem{EAST}
Xinyu Zhou, Cong Yao, He Wen, Yuzhi Wang, Shuchang Zhou, Weiran He, and Jiajun
  Liang.
\newblock East: an efficient and accurate scene text detector.
\newblock In {\em Proceedings of the IEEE conference on Computer Vision and
  Pattern Recognition}, pages 5551--5560, 2017.

\bibitem{GTR}
Xingyi Zhou, Tianwei Yin, Vladlen Koltun, and Philipp Kr{\"a}henb{\"u}hl.
\newblock Global tracking transformers.
\newblock In {\em Proceedings of the IEEE/CVF Conference on Computer Vision and
  Pattern Recognition}, pages 8771--8780, 2022.

\bibitem{FGFA}
Xizhou Zhu, Yujie Wang, Jifeng Dai, Lu Yuan, and Yichen Wei.
\newblock Flow-guided feature aggregation for video object detection.
\newblock In {\em ICCV}, pages 408--417, 2017.

\end{thebibliography}
}

\end{document}